\pdfoutput=1

\documentclass[11pt]{article}

\usepackage{acl}
\usepackage{graphicx}
\usepackage{booktabs}
\usepackage{multirow}
\usepackage{url}
\usepackage{times}
\usepackage{latexsym}
\usepackage{enumitem}
\usepackage{amsmath}
\usepackage{amsfonts}
\usepackage{marvosym}
\setenumerate[1]{itemsep=0pt,partopsep=0pt,parsep=\parskip,topsep=2pt}

\setitemize[1]{itemsep=0pt,partopsep=0pt,parsep=\parskip,topsep=2pt}

\setdescription{itemsep=0pt,partopsep=0pt,parsep=\parskip,topsep=2pt}
\usepackage[T1]{fontenc}

\usepackage[utf8]{inputenc}

\usepackage{microtype}

\title{Relation-Specific Attentions over Entity Mentions for\\ Enhanced Document-Level Relation Extraction}

\author{Jiaxin Yu$^{\dagger}$ \and Deqing Yang$^{\dagger}$ \Thanks {Corresponding author.} \and Shuyu Tian$^{\ddagger}$ \\
        School of Data Science, Fudan University, Shanghai 200433, China \\
        $^{\dagger}$\texttt{\{jiaxinyu20,yangdeqing\}@fudan.edu.cn}\\
        $^{\ddagger}$\texttt{sytian21@m.fudan.edu.cn}
        }

\begin{document}
\maketitle
\begin{abstract}
Compared with traditional sentence-level relation extraction, document-level relation extraction is a more challenging task where an entity in a document may be mentioned multiple times and associated with multiple relations. However, most methods of document-level relation extraction do not distinguish between mention-level features and entity-level features, and just apply simple pooling operation for aggregating mention-level features into entity-level features. As a result, the distinct semantics between the different mentions of an entity are overlooked. To address this problem, we propose \emph{RSMAN} in this paper which performs selective attentions over different entity mentions with respect to candidate relations. In this manner, the flexible and relation-specific representations of entities are obtained which indeed benefit relation classification. Our extensive experiments upon two benchmark datasets show that our RSMAN can bring significant improvements for some backbone models to achieve state-of-the-art performance, especially when an entity have multiple mentions in the document.\footnotemark[1]     

\end{abstract}

\footnotetext[1]{Our code and trained model are publicly available at \url{https://github.com/FDUyjx/RSMAN}.}

\section{Introduction}\label{intro}
Relation extraction (RE) is one important task of information extraction, aiming to detect the relations among entities in plain texts. Recently, many scholars have paid more attention to document-level RE \cite{sahu-etal-2019-inter,yao-etal-2019-docred} which aims to identify the relations of all entity pairs in a document, since it is more in demand than sentence-level RE in various real scenarios. In general, one document contains multiple entities and an entity may have multiple mentions across different sentences. Furthermore, one entity may be involved by multiple valid relations and different relations are expressed by different mentions of the same entity. As a result, document-level RE is more challenging than sentence-level RE.
\begin{figure}[t]
\centering
    \includegraphics[width=2.4in]{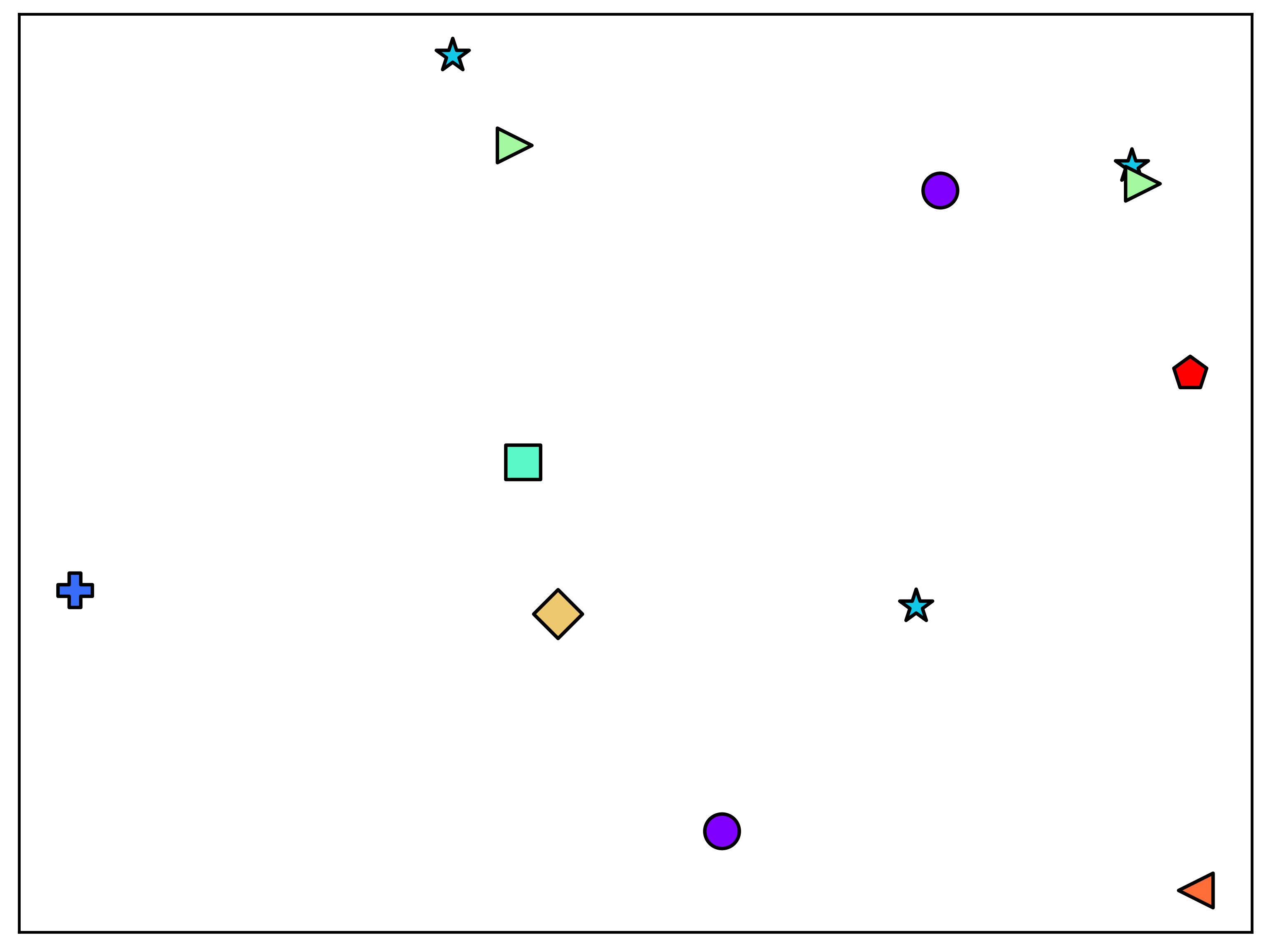}
    \caption{A \emph{t}-SNE visualization example from DocRED. Points of the same color and marker are different mentions' embeddings of an entity, which are encoded by BERT \cite{devlin-etal-2019-bert}.}
    \label{fig:tsne}
\end{figure}

A key step of existing document-level RE methods is to aggregate the information of different mentions of an entity (mention-level features) to obtain the entity's representation (entity-level feature) at first, since relation classification is generally achieved on entity level. To this end, 
previous RE models simply apply average pooling \cite{ye-etal-2020-coreferential,xu2021entity}, max pooling \cite{li-etal-2021-mrn}, or logsumexp pooling \cite{zhou2021document,zhang2021document}. Finally, a fixed representation is obtained for the given entity, which is then fed into the classifier for relation classification.

\begin{figure}[t]
    \includegraphics[width=3.0in]{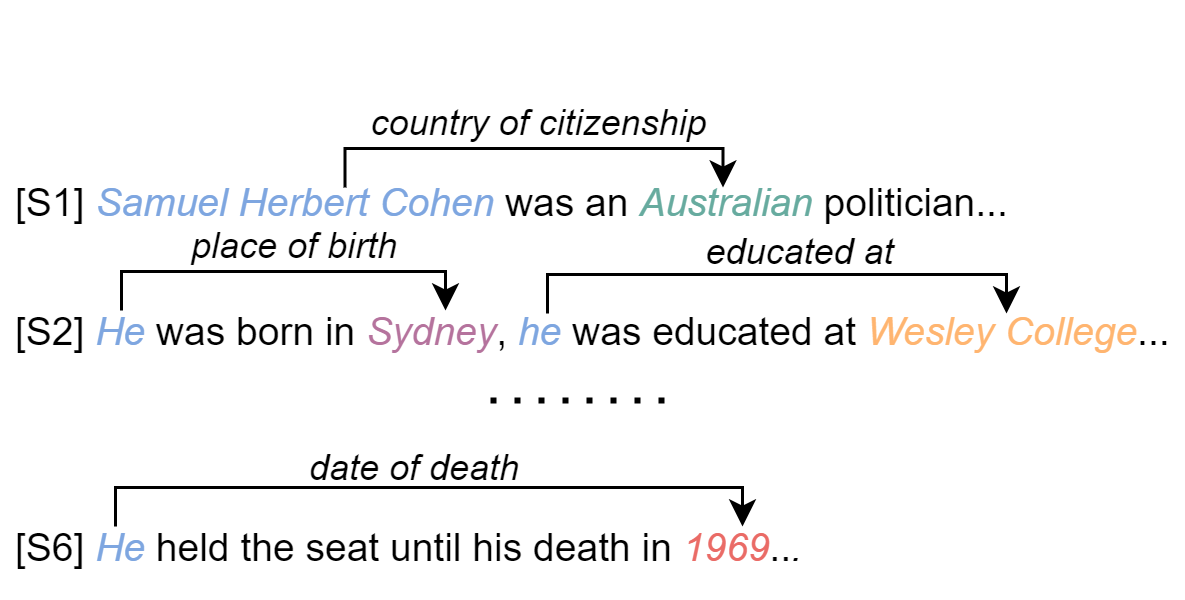}
    \caption{A toy example of multi-mention problem from the DocRED dataset. Based on coreference resolution, mentions belonging to the same entity are in the same color, and the relations are marked above the arrows.}
    \label{fig:casestudy}
\end{figure}

However, different mentions of an entity in a document may hold distinct semantics. A simple pooling operation of generating a fixed entity representation may confound the semantics of different mentions, and thus degrades the performance of relation classification when the entity is involved by multiple valid relations. We call such situation as \emph{multi-mention} problem in this paper. In Fig. \ref{fig:tsne}, we display the \emph{t}-SNE \cite{van2008visualizing} visualization of a toy example's mention embedding space to validate this problem. As the figure shows, different mentions' embeddings of an entity (marked by the same color) in a document are scattered over the whole embedding space, indicating that different mentions of an entity are not semantically adjacent. We further illustrate it by the toy example in Fig. \ref{fig:casestudy}, the first mention \textit{Samuel Herbert Cohen} of the person entity is more important for the classifier to identify the relation \textit{country of citizenship} between him and \textit{Australian}. But for extracting the relation \textit{place of birth}, the second mention \textit{He} should be considered more. It implies that different mentions should play different roles when extracting the different relations involving the same entity. In other words, different mentions function differently in different relation recognitions. 

Inspired by this intuition, we propose a novel \textbf{R}elation-\textbf{S}pecific \textbf{M}ention \textbf{A}ttention \textbf{N}etwork (\textbf{RSMAN}) to improve the model performance of document-level RE. 
In RSMAN, each relation's essential semantics is first encoded into a prototype representation. Then, the relevance weight (attention) between the prototype of a specific candidate relation and each mention's representation of the given entity is calculated. Based on these attentions, we get an attentive (weighted) sum of all mentions' representations as the entity's synthetic representation. In this manner, RSMAN enables the model to attend to the information of multiple mentions from different representation space when representing an entity, indicating that the entity's representation is flexible and relation-specific with respect to different candidate relations.

Our contributions in this paper can be summarized as follows:

\noindent 1. To the best of our knowledge, this is the first to consider different mentions' significance with respect to candidate relations on representing an entity to achieve document-level RE.

\noindent 2. We propose a novel RSMAN which can be used as a plug-in of a backbone RE model, to learn a relation-specific representation for a given entity which enhances the model's performance further.

\noindent 3. Our empirical results show that RSMAN can significantly promote some backbone models to achieve state-of-the-art (SOTA) RE performance, especially when an entity have multiple mentions in the document.

The rest of this paper is organized as follows. In Section 2, we briefly introduce some works related to our work. Then we introduce the proposed method in Section 3 and the experiment results in Section 4. At last, we conclude our work in Section 5.

\section{Related Work}\label{sec:rw}

Prior efforts on document-level RE mainly focused on representation learning and reasoning mechanism. \citet{yao-etal-2019-docred} employed four different sentence-level representation models to achieve document-level RE, including {CNN}, {LSTM}, {BiLSTM}, and {Context-Aware}. For more powerful representations, later work introduced pre-trained language models into their neural architectures \cite{ye-etal-2020-coreferential,zhou2021document,xu2021entity}. In particular, \citet{ye-etal-2020-coreferential} added a novel mention reference prediction task during pre-training and presented {CorefBERT} to capture the coreferential relations in contexts. \citet{zhou2021document} proposed {ATLOP} to learn an adjustable threshold and thus enhanced the entity pair's representation with localized context pooling. \citet{xu2021entity} defined various mention dependencies in a document and proposed {SSAN} to model entity structure for document-level RE. In addition, other work built various kinds of document graphs to model reasoning mechanism explicitly
\cite{nan-etal-2020-reasoning,zeng-etal-2020-double,wang-etal-2020-global}. For example, \citet{nan-etal-2020-reasoning} induced the latent document-level
graph and performed multi-hop reasoning on the induced latent structure. \citet{wang-etal-2020-global} constructed a global heterogeneous
graph and used a stacked R-GCN \cite{schlichtkrull2018modeling} to encode the document information. \citet{zeng-etal-2020-double} proposed {GAIN} to leverage both mention-level graph
and entity-level graph to infer relations between
entities. However they all ignore the multi-mention problem described in Sec. \ref{intro}.

\begin{figure*}[t]
\centering
    \includegraphics[width=4.7in]{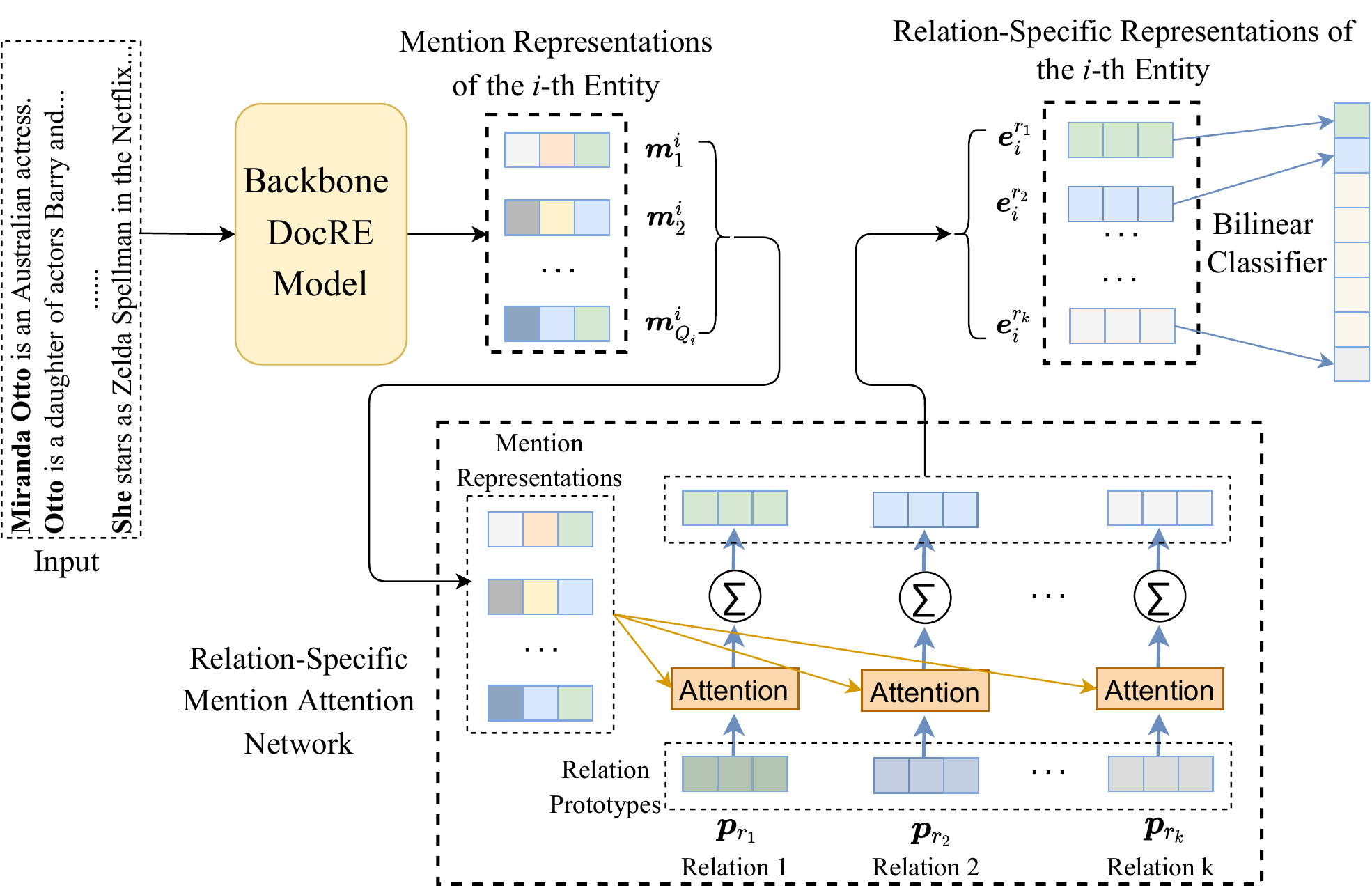}
    \caption{The overall work flow of RSMAN. The entity representations are obtained as the relation-specific weighted sum of mention representations.} \label{overview}
\end{figure*}

\section{Methodology}

At first, we formalize the task of document-level RE addressed in this paper as follows.


Suppose a document $\mathcal{D}$ mentions $P$ entities, denoted as $\mathcal{E}=\{{e_i}\}^{P}_{i=1}$, and the $i$-th entity $e_i$ has $Q_i$ mentions in $\mathcal{D}$, denoted as $\{{m^i_j}\}^{{Q_i}}_{j=1}$, the task of document-level RE is to extract a set of relational triples $\{(e_s,r,e_o)|e_s,e_o \in \mathcal{E},r \in \mathcal{R}\}$ where $\mathcal{R}$ is a pre-defined relation set.


\subsection{Backbone RE Model}


Suppose for each mention of $e_i$, its representation $\boldsymbol{m}^i_j$ is obtained by a model-specific method. Most of existing backbone models apply a certain pooling operation for all $\boldsymbol{m}^i_j$s to obtain $e_i$'s representation $\boldsymbol{e}_i$, such as the following average pooling,
\begin{equation}\label{en_co}
   \boldsymbol{e}_i=\frac{1}{Q_i}\sum_{j=1}^{Q_i}\boldsymbol{m}_j^i.
\end{equation}
As we claimed in Section \ref{intro}, $\boldsymbol{e}_i$ is a fixed representation which ignores that different mentions of $e_i$ play distinct roles when identifying the different relations involving $e_i$. 

Finally, given the subject entity's $\boldsymbol{e}_{s}$ and the object entity's representation $\boldsymbol{e}_{o}$, a bilinear classifier is often used to calculate the probability of relation $r$ involving these two entities as follows
\begin{equation}\label{bilinear}
   P(r|{e}_{s},{e}_{o}) = \sigma(\boldsymbol{e}_{s}^\top \boldsymbol{W}_r \boldsymbol{e}_{o}+b_r)
\end{equation}
where $\boldsymbol{W}_r \in \mathbb{R}^{d \times d}$ and $b_r \in \mathbb{R}$ are trainable model parameters specific to $r$, and $\sigma$ is Sigmoid activation. 

\subsection{Attentive Operations in RSMAN}
Our proposed RSMAN incorporates attentive mention-level features to generate flexible entity representations with respect to different candidate relations, and thus enhances the backbone model's performance. RSMAN's framework is shown in Fig. \ref{overview}, which acts as a plug-in of the backbone model. 

For each candidate relation $r$, its prototype representation $\boldsymbol{p}_r$ is first obtained through random initialization and is trainable during the training process. Then, we leverage $\boldsymbol{p}_r$ to calculate the semantic relevance between $r$ and each mention $m^i_j$ as follows,
\begin{equation}
   {s}^{r}_{ij} = g(\boldsymbol{p}_r,\boldsymbol{m}^i_j)
\end{equation}
where $g$ is a certain function to compute the similarity between two embeddings, which can be a simple dot-product or multi-layer perceptron (MLP) fed with the concatenation of two embeddings. Then, we feed all $s^{r}_{ij}$s of $e_i$ into a softmax function to get final attention weight
\begin{equation}
   {\alpha}^{r}_{ij} = \frac{\exp(s^{r}_{ij})}{{\sum^{Q_i}_{k=1}}\exp({s^{r}_{ik}})}.
\end{equation}

Since there is a necessity to consider all the mention information of the entity, we  use a weighted sum of all mention representations to obtain the relation-specific entity representation instead of using only one specific mention representation. We get $e_i$'s representation specific to $r$ as
\begin{equation}
   \boldsymbol{e}_i^r = \sum_{j=1}^{Q_i}{\alpha}^{r}_{ij}\boldsymbol{m}^i_j.
\end{equation}
Different to the fixed representation computed by Eq. \ref{en_co}, such $\boldsymbol{e}_i^r$ is a flexible embedding adaptive to different candidate relation $r$.

At last, we use this relation-specific entity representation to achieve relation classification by modifying Eq. \ref{bilinear} as
\begin{equation}
   P(r|{e}_{s},{e}_{o}) = \sigma({\boldsymbol{e}_{s}^{r}}^\top \boldsymbol{W}_r \boldsymbol{e}_{o}^{r}+b_r).
\end{equation}

\section{Experiments}
In this section, we introduce our experiments to justify our RSMAN, and provide insight into the experiment results.

\subsection{Datasets and Evaluation Metrics}
We conducted our experiments on two representative document-level RE datasets: DocRED \cite{yao-etal-2019-docred} and DWIE \cite{zaporojets2021dwie}, which are introduced in detail in Appendix \ref{sec:data}. We adopted F1 and Ign F1 as our evaluation metrics as \cite{yao-etal-2019-docred}, where Ign F1 is computed by excluding the common relation facts shared by the training, development (dev.) and test sets.

\subsection{Experimental Settings}
We use dot-product as the similarity scoring function for its computational efficiency, and before it we add a fully
connected layer to project the mention representations into the same embedding space with the prototype representations. All the additional parameters we introduce for RSMAN including the prototype representations is much fewer than the parameters of either the original bilinear classifier or the backbone model itself.

We took some stat-of-the-art models mentioned in Sec. \ref{sec:rw} as the baselines, i.e., \textbf{CNN} \cite{zeng2014relation}, \textbf{LSTM/BiLSTM} \cite{cai-etal-2016-bidirectional}, \textbf{Context-Aware} \cite{sorokin-gurevych-2017-context}, \textbf{CorefBERT} \cite{ye-etal-2020-coreferential}, \textbf{GAIN} \cite{zeng-etal-2020-double}, \textbf{SSAN} \cite{xu2021entity} and \textbf{ATLOP} \cite{zhou2021document}. We chose CorefBERT and SSAN as the backbone models in our framework due to their good performance and strong pluggability for our RSMAN. We did not consider GAIN and ATLOP as the backbone because they both leverage extra information besides entity representations. More setting details are shown in Appendix \ref{sec:details}.

\begin{table}[t]
  \centering
  \resizebox{\linewidth}{!}{
    \begin{tabular}{lccc}
    \toprule
    \multirow{2}[2]{*}{\textbf{Model}} & \textbf{Dev} &       & \textbf{Test} \\
          & \textbf{Ign F1} / \textbf{F1} &       & \textbf{Ign F1} / \textbf{F1} \\
    \midrule
    CNN*   & 37.65 / 47.73 &       & 34.65 / 46.14 \\
    LSTM*  & 40.86 / 51.77 &       & 40.81 / 52.60 \\
    BiLSTM* & 40.46 / 51.92 &       & 42.03 / 54.47 \\
    Context-Aware* & 42.06 / 53.05 &       & 45.37 / 56.58 \\
    CorefBERT & 57.18 / 61.42 &       & 61.71 / 66.59 \\
    GAIN*  & 58.63 / 62.55 &       & 62.37 / 67.57 \\
    SSAN  & 58.62 / 64.49 &       & 62.58 / 69.39 \\
    ATLOP* & 59.03 / 64.82 &       & 62.09 / 69.94 \\
    \midrule
    CorefBERT+RSMAN & 58.29 / 62.59 &       & 62.01 / 67.52 \\
    SSAN+RSMAN & \textbf{60.02} / \textbf{65.88} &       & \textbf{63.42} / \textbf{70.95} \\
    \bottomrule
    \end{tabular}}
  \vspace{-0.2cm}
    \caption{Performance (\%) comparisons on the dev. and test set of DWIE. The results with * are reported in \cite{ru-etal-2021-learning}. 
    }
  \label{dwie_res}%
\end{table}%

\begin{table}[t]
  \centering
  \resizebox{\linewidth}{!}{
    \begin{tabular}[!htb]
    {lccc}
    \toprule
    \multirow{2}[2]{*}{\textbf{Model}} & \textbf{Dev} &       & \textbf{Test} \\
          & \textbf{Ign F1} / \textbf{F1} &       & \textbf{Ign F1} / \textbf{F1} \\
    \midrule
    CorefBERT & 55.32 / 57.51 &       & 54.54 / 56.96 \\
    CorefBERT+RSMAN & \textbf{56.26} / \textbf{58.24} &       & \textbf{55.30} / \textbf{57.53} \\
    \midrule
    SSAN  & 56.68 / 58.95 &       & 56.06 / 58.41 \\
    SSAN+RSMAN & \textbf{57.22} / \textbf{59.25} &       & \textbf{57.02 / 59.29} \\
    \bottomrule
    \end{tabular}}
 \vspace{-0.2cm}
  \caption{Comparison on DocRED. The results of baselines are from their related papers. All test results are obtained by submitting to official Codalab\footnotemark[2].}
  \label{docred_res}%
   \vspace{-0.2cm}
\end{table}%
\footnotetext[2]{\url{https://competitions.codalab.org/competitions/20717}}

\subsection{Results and Analyses}\label{res and ana}
All the following results of our method were reported as the average scores of three runs. From the results on DWIE shown in Table \ref{dwie_res} we find that plugged with RSMAN, both CorefBERT and SSAN have significant improvements. Specifically, our RSMAN relatively improves CorefBERT's F1 by 1.9\% (dev. set) and 1.4\% (test set), and relatively improves SSAN's F1 by 1.84\% dev F1 (dev. set) and 2.25\% (test set), respectively. The consistent improvements verify the effectiveness of leveraging attentive mention-level features to learn relation-specific entity representations. What's more, the positive effects on different backbone models show good generalization performance of our RSMAN. Overall, SSAN+RSMAN achieves 63.42\% Ign F1 and 70.95\% F1 on the test set, outperforming all the baselines apparently.

For simplicity, we only display the results on DocRED of CorefBERT and SSAN plugged with RSMAN in Table \ref{docred_res}. It shows that RSMAN also brings relative improvements of 1.39\% Ign F1 and 1.00\% F1 on the test set for CorefBERT, along with relative improvements of 1.71\% Ign F1 and 1.51\% F1 for SSAN. It is worth noting that the performance improvements on DocRED are relatively less significant than that on DWIE. Through our statistics, we found that the average number of mentions per entity in DocRED is only 1.34, while it is 1.98 in DWIE. Besides, only 18.49\% of entities in DocRED have multiple mentions, much less than 33.59\% in DWIE. It implies that our RSMAN is more effective on the entities with multiple mentions, which are more common and challenging in many real scenarios of document-level RE.

\begin{figure}[t]
\centering
\vspace{-0.1cm}
    \includegraphics[width=2.3in]{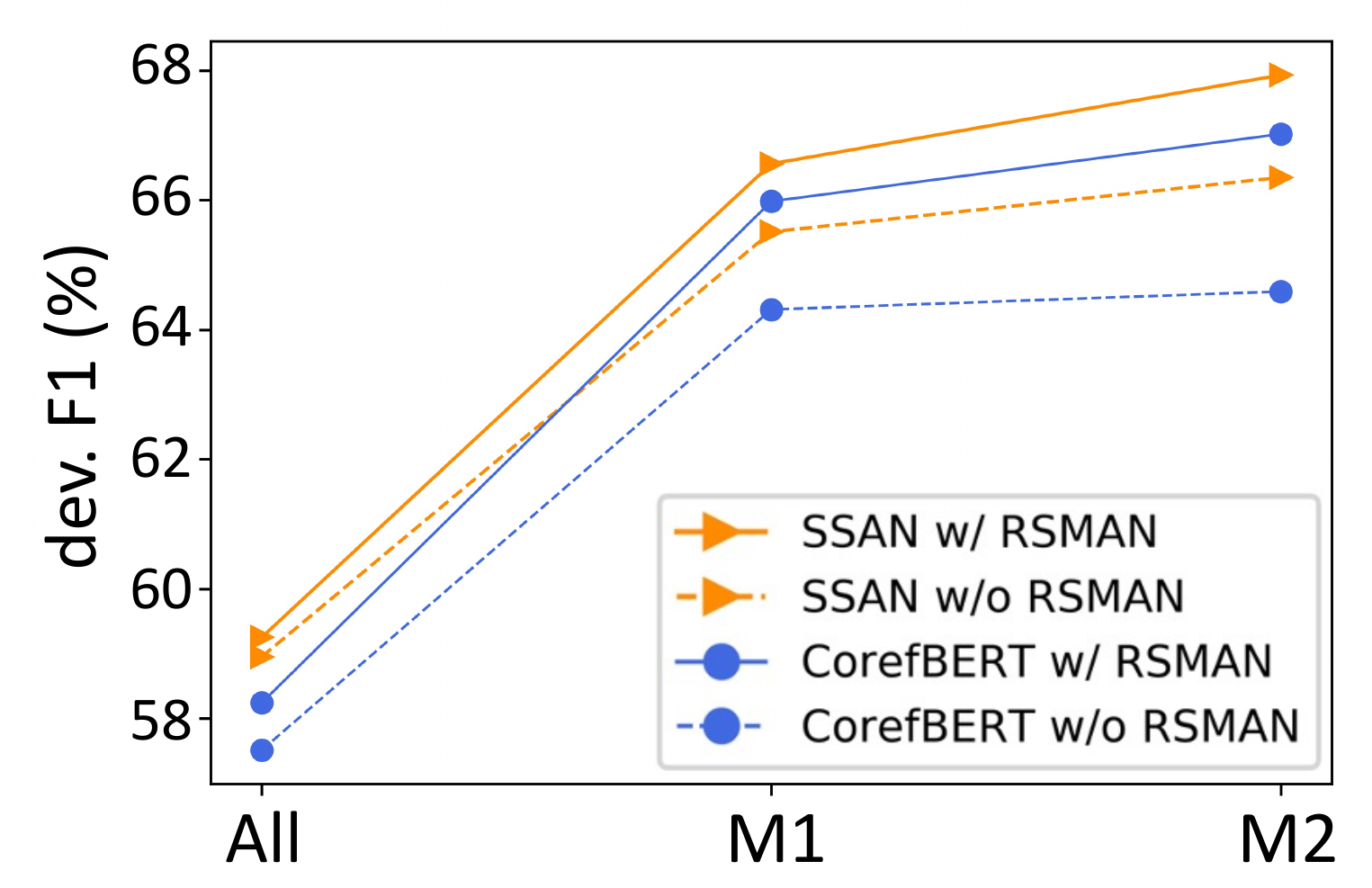}
    \vspace{-0.2cm}
    \caption{F1 variations on three subsets reconstructed from the development set of DocRED.} \label{relation instance}
    \vspace{-0.2cm}
\end{figure}

\begin{figure}[t]
\centering
    \includegraphics[width=3.4in]{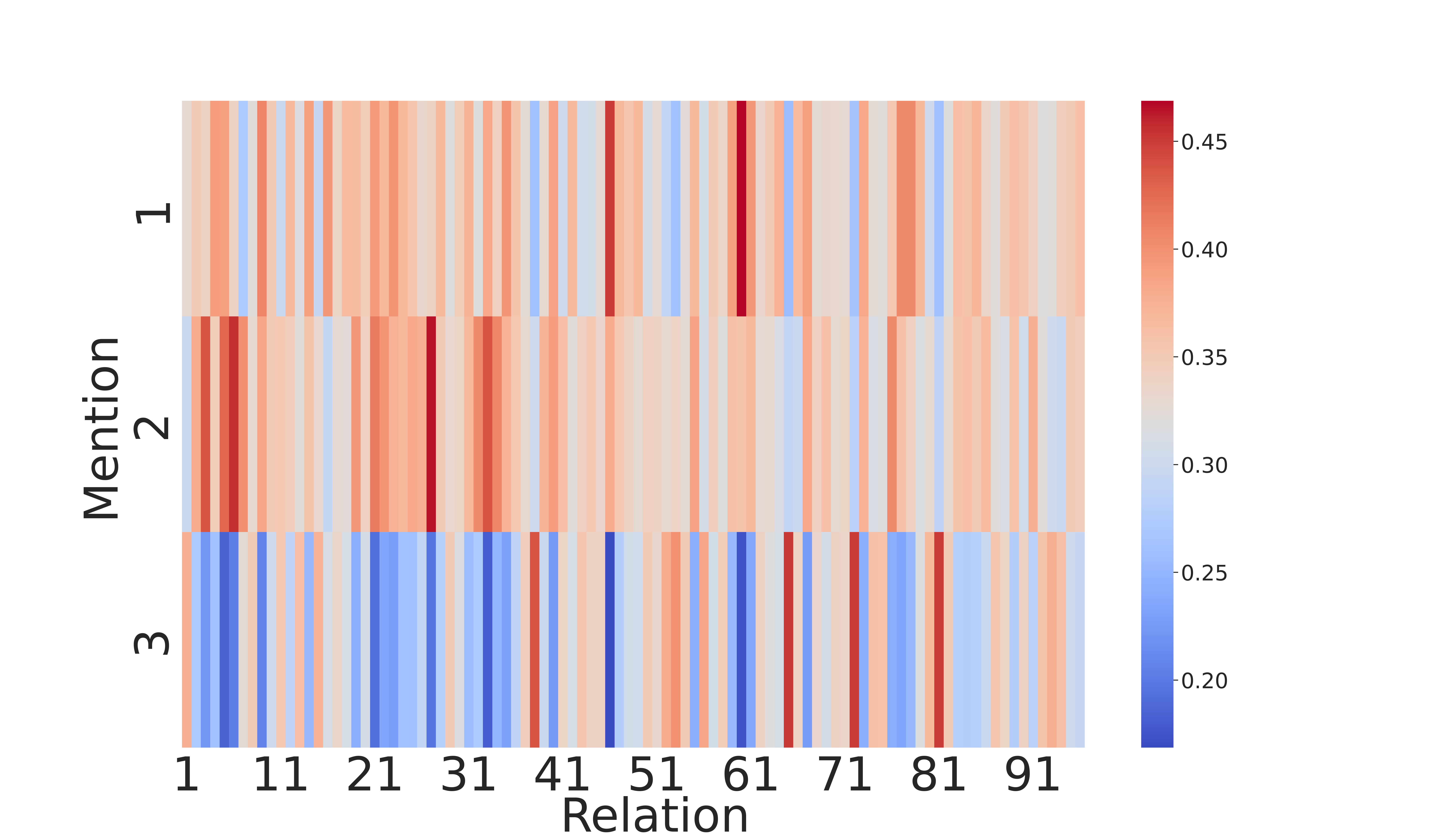}
    \vspace{-0.5cm}
    \caption{Visualization on relation attentions to different mentions of a given entity.}
    \label{heatmap}
  \vspace{-0.2cm}
\end{figure}

\subsection{Effect Analysis for Mention Number}
To confirm our conjecture mentioned before, we investigated the effect of mention number through further experiments. We first reconstructed the relation instances in DocRED's dev. set and obtained three different subsets: the first one contains all instances (All), another one contains either subject or object argument having more than one mention (M1), and the rest one contains either subject or object argument having more than two mentions (M2). We don't consider M3 or higher because they have very few instances limited by the dataset. Then, we evaluated CorefBERT and SSAN with or without RSMAN upon the three subsets.

From Fig. \ref{relation instance}, we find that the F1s of all compared methods increase from All to M2. It indicates that multiple mentions can provide more information for the models to capture the entity semantics, resulting in more precise RE results. Furthermore, the performance gains of plugging RSMAN into the two backbone models also increase as the mention number per entity increases. It shows that our RSMAN can bring more significant performance boosts for the backbone model when the entities of the relation instances have more mentions in a document. These results justify that RSMAN has more potential for extracting relations based on the entities with more mentions.

\subsection{Case Study}
To explore how RSMAN attends to different mentions' information of an entity, we collected all relations' normalized attentions for an entity's mentions in RSMAN. Fig. \ref{heatmap} is the heatmap of attentions for a specific entity, from which we observe that the distribution of relation attentions varies greatly among different mentions. Besides, according to the high attention of a given relation, we can capture which mention of the entity well expresses this relation's semantics. This map also confirms the implication of Fig. \ref{fig:tsne} that different mentions of an entity contain distinct semantics. Therefore, the attentive aggregation of all mention-level features in RSMAN is more appropriate for enhanced RE than the common pooling operations.

\section{Conclusion}

In this paper, we focus on the multi-mention problem in document-level RE and propose RSMAN to address it. Our experiment results
demonstrate RSMAN's effectiveness especially on the scenario of multi-mention situation. In the future, we plan to adapt RSMAN to more document-level RE models.

\section*{Acknowledgements}
This paper was supported by Shanghai Science and Technology Innovation Action Plan No.21511100401, and AECC Sichuan Gas Turbine Establishment (No.GJCZ-2019-0070), Mianyang Sichuan, China. We sincerely thank all reviewers for their valuable comments to improve our work.

\bibliography{anthology,custom}
\bibliographystyle{acl_natbib}

\appendix
\section{Datasets}\label{sec:data}
DocRED is a large-scale human-annotated dataset for document-level RE. DWIE is a dataset for document-level multi-task information extraction which combines four main sub-tasks and in our work we only used the dataset for document-level relation extraction. 
We preprocessed DWIE dataset and adopted the same dataset partition as \cite{ru-etal-2021-learning}. More statistical information is detailed in Table \ref{tab:statistics}.
\begin{table}[htbp]
  \centering
    \begin{tabular}{lcc}
    \toprule
    Statistics & \multicolumn{1}{l}{DWIE} & \multicolumn{1}{l}{ DocRED} \\
    \midrule
    \# Train & 602   & 3053 \\
    \# Dev & 98    & 1000 \\
    \# Test & 99    & 1000 \\
    \# Relations & 65    & 96 \\
    \# Relation facts & 19493 & 56354 \\
    Avg.\# mentions per Ent. & 1.98  & 1.34 \\
    \bottomrule
    \end{tabular}%
    \caption{Statistics of the two datasets.}
  \label{tab:statistics}%
\end{table}%

\begin{table}[htbp]
  \centering
    \begin{tabular}{lcc}
    \toprule
    Hyper-parameter & \multicolumn{1}{l}{DWIE} & \multicolumn{1}{l}{ DocRED} \\
    \midrule
    Batch size & 4     & 8 \\
    Learning rate & {3e-5} & {5e-5} \\
    Epoch & 40    & 60 \\
    Gradient clipping & 1     & 1 \\
    Warmup ratio & 0.1   & 0.1 \\
    \bottomrule
    \end{tabular}%
    \caption{Hyper-parameter settings for our experiments on the two datasets.}
  \label{parameter}%
\end{table}%

\section{Implementation Details}
\label{sec:details}

In this appendix, we introduce more details of our experimental settings. We implemented our RSMAN with PyTorch and trained it with an NVIDIA GeForce RTX 3090 GPU. In addition, we adopted AdamW \cite{loshchilov2018decoupled} as our optimizer and used learning rate linear schedule with warming up based on Huggingface’s Transformers \cite{wolf2019huggingface}.
The hyper-parameter settings of our experiments on the two datasets are listed in Table \ref{parameter}, which were decided through our tuning studies.

\end{document}